\documentclass{article}

\usepackage{arxiv}

\usepackage[utf8]{inputenc} 
\usepackage[T1]{fontenc}    
\usepackage{hyperref}       
\usepackage{url}            
\usepackage{booktabs}       
\usepackage{amsfonts}       
\usepackage{nicefrac}       
\usepackage{microtype}      
\usepackage{lipsum}

\usepackage{bm}
\usepackage{amsmath}
\usepackage{amssymb}
\usepackage{amsfonts}
\usepackage{graphicx}
\def\vec#1{\mathbf{#1}}
\usepackage{algorithm}
\usepackage{algorithmic}
\usepackage{comment}

\DeclareMathOperator*{\argmin}{arg\,min}

\title{Meta-learning of Physics-informed Neural Networks for Efficiently Solving Newly Given PDEs}

\author{
  Tomoharu Iwata\\
  NTT Communication Science Laboratories, NTT Corporation\\  
  \AND
  Yusuke Tanaka\\
  NTT Communication Science Laboratories, NTT Corporation\\  
  \AND
  Naonori Ueda\\
  NTT Communication Science Laboratories, NTT Corporation\\
}
\date{}

\renewcommand{\shorttitle}{Meta-learning of Physics-informed Neural Networks}

\begin{document}
\maketitle

\begin{abstract}
We propose a neural network-based meta-learning method to efficiently solve partial differential equation (PDE) problems. The proposed method is designed to meta-learn how to solve a wide variety of PDE problems, and uses the knowledge for solving newly given PDE problems. We encode a PDE problem into a problem representation using neural networks, where governing equations are represented by coefficients of a polynomial function of partial derivatives, and boundary conditions are represented by a set of point-condition pairs. We use the problem representation as an input of a neural network for predicting solutions, which enables us to efficiently predict problem-specific solutions by the forwarding process of the neural network without updating model parameters. To train our model, we minimize the expected error when adapted to a PDE problem based on the physics-informed neural network framework, by which we can evaluate the error even when solutions are unknown. We demonstrate that our proposed method outperforms existing methods in predicting solutions of PDE problems.
\end{abstract}

\section{Introduction}

Partial differential equations (PDEs) have been used in many areas of science and engineering
for modeling physical~\cite{lin1988mathematics,courant2008methods}, biological~\cite{jones2009differential},
and finaicial~\cite{duffy2013finite} systems.
The finite element method (FEM)~\cite{turner1956stiffness}
is the standard method for numerically solving PDEs.
Recently, physics-informed neural networks (PINNs)~\cite{raissi2019physics,karniadakis2021physics}
have been attracting attention as a machine learning approach for solving PDEs.
However, both methods are computationally expensive.
In addition, they solve PDE problems from scratch even when
similar PDE problems have been solved before.

\begin{figure}[t!]
  \centering
  \includegraphics[width=43em]{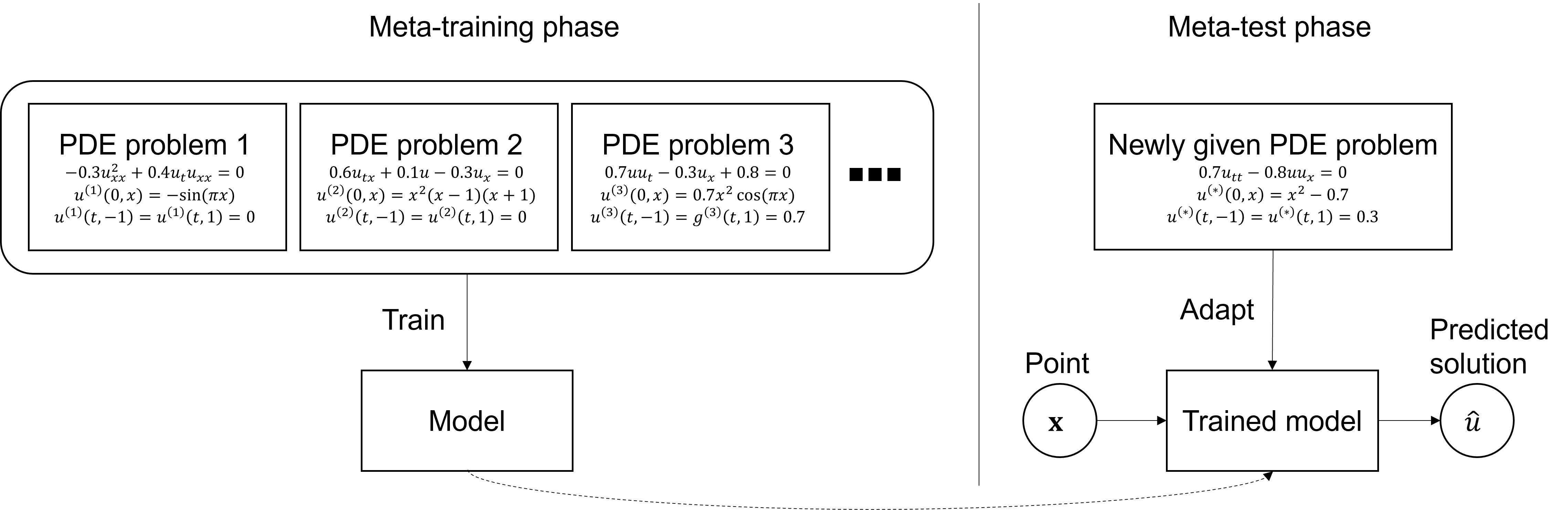}
  \caption{Our meta-learning framework. Left) In the meta-training phase, our neural network-based model is trained using many PDE problems with various governing equations and boundary conditions. Right) In the meta-test phase, we are given new PDE problems. The trained model is adapted to each PDE problem. The adapted model outputs approximated problem-specific solution $\hat{u}$ at point $\vec{x}$.}
  \label{fig:framework}
\end{figure}

In this paper, we propose a meta-learning method for efficiently solving PDEs.
The proposed method learns how to solve PDEs with our neural network-based model using many PDE problems,
and uses the knowledge for solving newly given PDE problems.
Figure~\ref{fig:framework} shows the framework of the proposed method.
Our model takes a PDE problem and a point as input,
and outputs a predicted solution at the point of the PDE problem.
The neural networks are shared across problems,
by which we can accumulate the knowledge of solving various problems in their parameters.
Although meta-learning methods for PINNs have been
proposed~\cite{de2021hyperpinn,qin2022meta,huang2022meta,pan2023neural},
these existing methods transfer knowledge within
a PDE problem with different physical parameters and/or different boundary conditions.
On the other hand, the proposed method can transfer knowledge across different PDE problems
if their governing equations are formulated as polynomial functions of partial derivatives.
  
In our model, we encode a PDE problem by neural networks.
Each PDE problem consists of a governing equation and boundary conditions.
We assume PDE problems with a governing equation defined by a polynomial function of partial derivatives,
and Dirichlet boundary conditions.
Many real-world PDE problems satisfy this assumption.
Coefficients of the polynomial are used as the representation of a governing equation,
by which we can encode a governing equation by a fixed size of a vector.
Dirichlet boundary conditions are represented by a set of pairs of points and their conditions,
where points are randomly generated on the boundaries.
Using such sets, we can encode boundary conditions flexibly,
even when boundary shapes and/or conditions differ across problems.
The set is transformed into a fixed size of a boundary representation
vector using a permutation invariant neural network.
A problem representation is obtained by a neural network
from the governing equation coefficients and boundary representation.
By feeding the problem representation and a point into a neural network,
we can output a problem-specific solution.
When a PDE problem is newly given, our model can efficiently predict its solution by
only the forwarding process of the neural networks without updating parameters.

To train our model,
we minimize the expected error of predicted solutions when adapted to each PDE problem.
The error is calculated using the PINN framework,
which enables us to evaluate the error using governing equations and
boundary conditions even when their solutions are unknown.
It is essential for efficient meta-learning
since obtaining solutions for many problems is very expensive to compute.
The expectation is approximated by the Monte Carlo method,
where PDE problems are randomly generated.
In the proposed method, the meta-training data can be generated unlimitedly
since we do not need any observational data.
It is preferable since the performance of neural networks generally improves
as the number of training data increases~\cite{kaplan2020scaling}.
The meta-learned model can be finetuned to a PDE problem,
by which we can improve the approximation performance of the solution.

The main contributions of this paper are as follows:
1) We present a meta-learning method for solving a broad class of PDE problems
based on PINNs, where governing equations and boundary conditions are different across problems.
2) We propose a neural network-based model that takes a PDE problem and a point as input
and outputs its solution.
3) We experimentally demonstrate that the proposed method can predict solutions
of newly given PDE problems in about a second, and its performance is better than the existing methods.

\section{Related work}

Several meta-learning methods for PINNs have been proposed,
such as Meta-PDE~\cite{qin2022meta}, HyperPINN~\cite{de2021hyperpinn}, and Meta-auto-decoders~\cite{huang2022meta}.
Meta-PDE~\cite{qin2022meta} meta-learns PINNs using model-agnostic meta-learning~\cite{finn2017model},
where initial neural network parameters are trained
such that it achieves good performance with gradient descent updates.
Since backpropagation through such procedures is costly in terms of memory,
the total number of updates must be kept small.
It is difficult to adapt to a wide range of PDE problems
with a small number of updates from a single initialization. 
HyperPINN~\cite{de2021hyperpinn}
is a hyper-network that generates a PINN given a PDE problem.
Meta-auto-decoders~\cite{huang2022meta} implicitly encode PDE problems into vectors,
and feed the encoded vectors to a PINN to predict their solutions.
These existing methods share knowledge
across a narrow range of PDE problems,
e.g., the Burgers' equation with different parameters with a fixed boundary condition.
In contrast, the proposed method shares the knowledge
across different classes of PDE problems by considering a general form of governing equations
and representing boundary conditions by a set of boundary points and their solutions.
When PDE problems are encoded implicitly as in Meta-auto-decoders,
the required memory linearly increases as the number of training PDE problems increases,
which prohibits to use a huge amount of PDE problems for training.
On the other hand, since the proposed method explicitly encodes PDE problems using neural networks,
the memory cost does not depend on the number of training PDE problems,
and we can use an unlimited number of PDE problems for training.

The proposed method uses an encoder-decoder style meta-learning~\cite{xu2020metafun},
or in-context learning~\cite{dong2023survey},
approach as with neural processes~\cite{garnelo2018conditional},
where neural networks are used for encoding problem (task) representations and for decoding solutions.
This approach approximates problem adaptation with neural networks,
and does not require gradient descent iterations for problem adaptation,
which enables us to perform meta-learning and inference efficiently.

\section{Preliminaries}

\subsection{PDE problem}
\label{sec:pde}

We consider the following parametric PDE problem on domain $\Omega^{(s)}\subset\mathbb{R}^{D}$
with boundary $\partial\Omega^{(s)}$,
\begin{align}
  \mathcal{F}^{(s)}[u^{(s)}(\vec{x})]=0, \quad \vec{x}\in\Omega^{(s)},
  \qquad
  u^{(s)}(\vec{x})=g^{(s)}(\vec{x}), \quad \vec{x}\in\partial\Omega^{(s)},
\end{align}
where $s$ is the index of the PDE problem,
$\mathcal{F}^{(s)}$ is the governing equation operator
that involve $u$ and partial derivatives of $u$
with respect to point $\vec{x}$,
$u^{(s)}(\vec{x})$ is the hidden solution,
and $g^{(s)}(\vec{x})$ is the Dirichlet boundary condition.
Each PDE problem has its own
governing equation,
boundary condition,
domain,
boundary,
and hidden solution.
For example, 
in the case of the Burgers' equation with point $\vec{x}=(t,x)$,
the governing equation is $\mathcal{F}^{(s)}=u_{t}+uu_{x}-\alpha u_{xx}$,
and the boundary condition can be
$g^{(s)}(\vec{x})=-\sin(\pi x)$ if $t=0$,
and 
$g^{(s)}(\vec{x})=0$ if $x=-1$ or $x=1$.

\subsection{Physics-informed neural networks}

PINNs approximate the solution of a PDE problem
using neural network $\hat{u}^{(s)}(\vec{x};\bm{\theta})$.
Here, $\bm{\theta}$ is the parameters,
and the neural network takes point $\vec{x}$ as input, and outputs the predicted solution. 
The parameters are trained by minimizing the
following objective function:
\begin{align}
  \hat{\bm{\theta}}=\argmin_{\bm{\theta}} \frac{1}{N_{\mathrm{f}}}\sum_{n=1}^{N_{\mathrm{f}}}
  \parallel \mathcal{F}^{(s)}\big(\hat{u}^{(s)}(\vec{x}_{\mathrm{f}n};\bm{\theta})\big)\parallel^{2}
  +
  \frac{1}{N_{\mathrm{g}}}\sum_{n=1}^{N_{\mathrm{g}}}
  \parallel g(\vec{x}_{\mathrm{g}n})-\hat{u}^{(s)}(\vec{x}_{\mathrm{g}n};\bm{\theta})\parallel^{2},
      \label{eq:pinn_loss}
\end{align}
where $\{\vec{x}_{\mathrm{f}n}\}_{n=1}^{N_{\mathrm{f}}}$ is a set of randomly sampled points in domain $\Omega^{(s)}$,
and $\{\vec{x}_{\mathrm{g}n}\}_{n=1}^{N_{\mathrm{g}}}$ is a set of randomly sampled points on boundary $\partial\Omega^{(s)}$.
The first and second terms enforce to satysify governing equation $\mathcal{F}^{(s)}$
and boundary condition $g^{(s)}$, respectively.
We call the objective function in Eq.~(\ref{eq:pinn_loss}) the PINN error.

\section{Proposed method}
\subsection{Problem formulation}

We consider parametric PDE problems
described in Section~\ref{sec:pde}.
Governing equations $\mathcal{F}^{(s)}$
for all problems are
assumed to be a polynomial function of partial derivatives
with a constant driving term,
where the maximum degree of the polynomial is $C$,
and the maximum order of the derivatives is $J$.
In our experiments, we consider $D=2$, $C=2$, and $J=2$.
Many widely-used PDEs are included in this PDE class,
such as
Burgers' equation, $u_{t}+uu_{x}-\alpha u_{xx}=0$, 
Fisher's equation, $u_{t}-\alpha_{1}u_{xx}-\alpha_{2}u(1-u)=0$,
Hunter-Saxton equation, $u_{tx}+u_{x}u_{xx}-\alpha u_{x}^{2}=0$,
heat equation, $u_{t}-u_{xx}=0$,
Laplace's equation, $u_{xx}+u_{yy}=0$, and
wave equation, $u_{tt}-\alpha u_{xx}=0$.
Unlike governing equations $\mathcal{F}^{(s)}$,
boundary condition $g^{(s)}$
is difficult to assume a specific functional form in real-world PDE problems.
Therefore, we do not assume such assumptions on the boundary condition.
Our aim is to predict solutions efficiently 
of newly given PDE problems
with a neural network by meta-learning from various PDEs.

\subsection{Model}

\begin{figure}[t]
  \centering
  \includegraphics[width=38em]{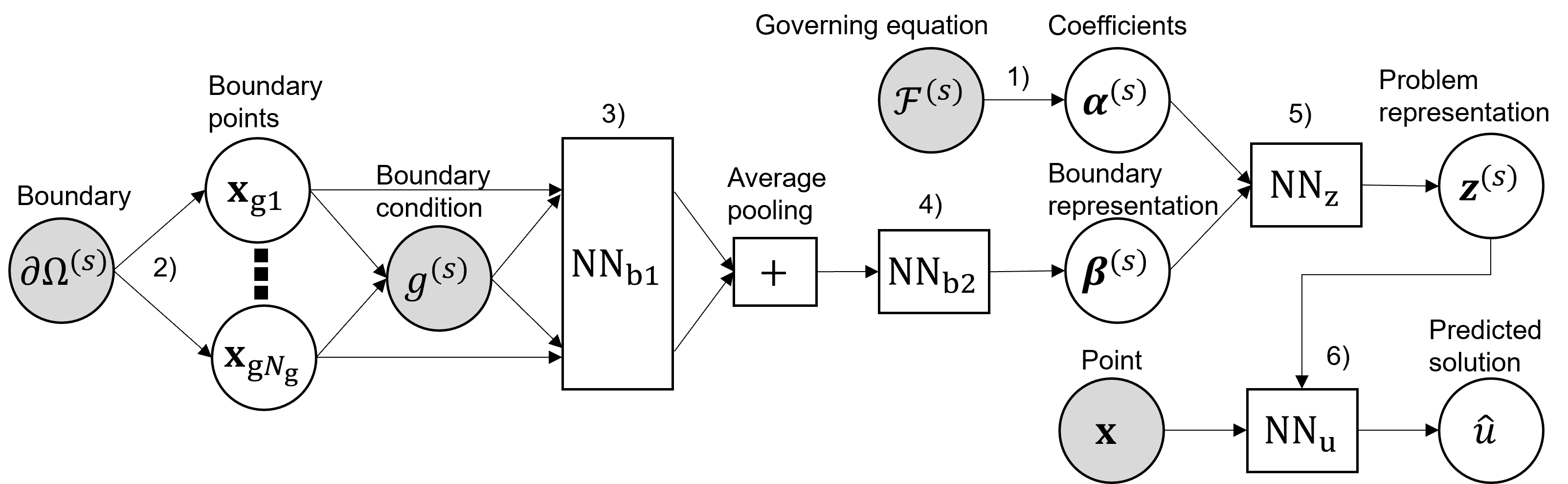}
  \caption{Our model outputs predicted solution $\hat{u}$ at point $\vec{x}$ given governing equation $\mathcal{F}^{(s)}$, boundary $\partial\Omega$, and boundary condition $g^{(s)}$.
    1) Governing equation $\mathcal{F}^{(s)}$ is represented by its coefficients $\bm{\alpha}^{(s)}$. 2) Points at boundary $\partial\Omega^{(s)}$ are randomly generated. 3) Each pair of boundary point $\vec{x}_{\mathrm{g}n}$ and its condition $g^{(s)}(\vec{x}_{\mathrm{g}n})$ is transformed into a vector by neural network $\mathrm{NN_{b1}}$. 4) These vectors are averaged, and then transformed to boundary representation $\bm{\beta}^{(s)}$ by neural network $\mathrm{NN_{b2}}$.
    5) Coefficients $\bm{\alpha}^{(s)}$ and boundary representation $\bm{\beta}^{(s)}$ are transformed into problem representation $\vec{z}^{(s)}$ by neural network $\mathrm{NN_{z}}$. 6) Using the problem representation, solution $\hat{u}$ at point $\vec{x}$ is predicted by neural network $\mathrm{NN_{u}}$. Shaded circles represent given information. Unshared circles represent variables calculated by our model. Rectangles represent operations in our model.}
  \label{fig:model}
\end{figure}

Our neural network-based model outputs
prediction of the problem-specific solution $\hat{u}$
at point $\vec{x}$ that is adapted to the given PDE problem with governing equation $\mathcal{F}^{(s)}$, 
boundary $\partial\Omega^{(s)}$, and boundary condition $g^{(s)}$.
Figure~\ref{fig:model} illustrates our model.

Since we assume that
the governing equation is 
a $C$th-degree polynomial of $J$th-order partial derivatives
with $D$-dimensional point,
it can be written in a general form of the polynomial.
For example, in the case with $D=2$ with point $\vec{x}=(t,x)$,
the general form is given by
\begin{align}
  \mathcal{F}^{(s)}=\sum_{c_{1}+c_{2}+c_{3}+c_{4}+c_{5}+c_{6}\in\{0,1,\dots,C\}}
  \alpha^{(s)}_{c_{1}c_{2}c_{3}c_{4}c_{5}c_{6}}u^{c_{1}}u_{t}^{c_{2}}u_{x}^{c_{3}}u_{tt}^{c_{4}}u_{tx}^{c_{5}}u_{xx}^{c_{6}},
\end{align}
where $\alpha^{(s)}_{c_{1}c_{2}c_{3}c_{4}c_{5}c_{6}}\in\mathbb{R}$ is the coefficient,
and the degree for each term is $C$ at most.
Let $\bm{\alpha}^{(s)}$ be a vector of coefficients $\{\alpha^{(s)}_{c}\}$.
The dimension of $\bm{\alpha}^{(s)}$ is $\binom{\binom{D+J}{J}+C}{C}$,
where $\binom{D+J}{J}$ is the number of partial derivatives $\{u,u_{t},u_{x},\dots\}$.
Governing equations with a lower degree of polynomial, lower order of partial derivatives,
or lower dimensional points can be represented by setting zero to the corresponding coefficients.
Our model uses the coefficients $\bm{\alpha}^{(s)}$
as the governing equation representation.

We introduce boundary condition $g^{(s)}$ into our model
by representing it by a set of the boundary condition data
at randomly sampled points.
In particular,
we randomly generate $N_{\mathrm{g}}$ points at boundary $\partial\Omega^{(s)}$,
$\vec{X}_{\mathrm{g}}=\{\vec{x}_{\mathrm{g}n}\}_{n=1}^{N_{\mathrm{g}}}$,
and evaluate the solution at the boundary points,
$\{g^{(s)}(\vec{x}_{\mathrm{g}n})\}_{n=1}^{N_{\mathrm{g}}}$.
The set of pairs
$\{(\vec{x}_{\mathrm{g}n},g^{(s)}(\vec{x}_{\mathrm{g}n}))\}_{n=1}^{N_{\mathrm{g}}}$
are transformed
into boundary representation $\bm{\beta}^{(s)}$
using the following permutation invariant neural network~\cite{zaheer2017deep}:
\begin{align}
  \bm{\beta}^{(s)}=
  \mathrm{NN_{b2}}\left(\frac{1}{N_{\mathrm{g}}}\sum_{n=1}^{N_{\mathrm{g}}}\mathrm{NN_{b1}}\left(\vec{x}_{\mathrm{g}n},g^{(s)}(\vec{x}_{\mathrm{g}n})\right)\right),
  \label{eq:b}
\end{align}
where 
$\mathrm{NN_{b1}}$ and $\mathrm{NN_{b2}}$ are feed-forward neural networks.
The output of the neural network in Eq.~(\ref{eq:b}) is invariant
even when the elements are permuted,
which is desirable since the order of the randomly sampled points should not depend on the boundary representation.
In addition, the neural network can take any number of points, $N_{\mathrm{g}}$, as input.
With sampled points at problem-specific boundary $\partial\Omega^{(s)}$,
the boundary representation can consider not only the solution at the boundary
but also the shape of the boundary.

Problem representation $\vec{z}^{(s)}$ is obtained by
a neural network from coefficients $\bm{\alpha}^{(s)}$ and boundary representation $\bm{\beta}^{(s)}$ as follows:
\begin{align}
  \vec{z}^{(s)}= \mathrm{NN_{z}}(\bm{\alpha}^{(s)},\bm{\beta}^{(s)}),
  \label{eq:z}
\end{align}
where $\mathrm{NN_{z}}$ is a feed-forward neural network.

Solution $\hat{u}$ at point $\vec{x}$ of PDE problem $s$
is predicted by inputting its problem representation $\vec{z}^{(s)}$ to a neural network as follows:
\begin{align}
  \hat{u}\left(\vec{x},\mathcal{F}^{(s)},g^{(s)},\vec{X}_{\mathrm{g}};\bm{\Theta}\right)=\mathrm{NN_{u}}(\vec{x},\vec{z}^{(s)}),
\end{align}
where $\mathrm{NN_{u}}$ is a feed-forward neural network,
and $\bm{\Theta}$ is unknown parameters in our model to be trained,
which are the parameters in neural networks $\mathrm{NN_{b1}}$,
$\mathrm{NN_{b2}}$, $\mathrm{NN_{z}}$, and $\mathrm{NN_{u}}$.
All of the parameters are shared across different PDE problems,
by which we can learn how to solve various PDE problems and use the knowledge for newly given PDE problems.
By taking governing problem representation $\vec{z}^{(s)}$ as input,
our model can output a problem-specific solution
that depends on given governing equation $\mathcal{F}^{(s)}$, boundary $\partial\Omega^{(s)}$, and boundary condition $g^{(s)}$.

\subsection{Meta-learning}
\label{sec:meta-learning}

We estimate unknown parameters $\bm{\Theta}$ by minimizing
the expected PINN error over PDE problems as follows:
\begin{align}
  \hat{\bm{\Theta}}=\argmin_{\bm{\Theta}}
  \mathbb{E}_{s}
  \Bigg[
    \mathbb{E}_{\vec{X}_{\mathrm{g}}\in\partial\Omega^{{(s)}^{N_{\mathrm{g}}}}}
    \bigg[
      &
      \mathbb{E}_{\vec{x}_{\mathrm{f}}\in\Omega^{(s)}}
      \Big[
        \parallel 
        \mathcal{F}^{(s)}\left(\hat{u}^{(s)}(\vec{x}_{\mathrm{f}},\mathcal{F}^{(s)},g^{(s)},\vec{X}_{\mathrm{g}})\right)
        \parallel^{2}
        \Big]
      \nonumber\\
    &+
      \frac{1}{N_{\mathrm{g}}}
        \sum_{n=1}^{N_{\mathrm{g}}}
        \parallel 
        g(\vec{x}_{\mathrm{g}n})-
        \hat{u}^{(s)}(\vec{x}_{\mathrm{g}n},\mathcal{F}^{(s)},g^{(s)},\vec{X}_{\mathrm{g}})
        \parallel^{2}
        \bigg]
    \Bigg],
  \label{eq:theta}
\end{align}
where $\mathbb{E}_{s}$ is the expectation over PDE problems,
$\mathbb{E}_{\vec{X}_{\mathrm{g}}\in\partial\Omega^{{(s)}^{N_{\mathrm{g}}}}}$
is the expectation over sets of boundary points $\vec{X}_{\mathrm{g}}$
at boundary $\partial\Omega^{(s)}$,
$\mathbb{E}_{\vec{x}_{\mathrm{f}}\in\Omega^{(s)}}$
is the expectation over points $\vec{x}_{\mathrm{f}}$ in domain $\Omega^{(s)}$,
the first term is the discrepancy between 
governing equation $\mathcal{F}^{(s)}$
and our model $\hat{u}$ at point $\vec{x}_{\mathrm{f}}$,
and the second term is the discrepancy between
boundary condition $g^{(s)}$ and
the prediction by our model $\hat{u}$ at boundary points $\vec{X}_{\mathrm{g}}$.
Partial differentiation $\mathcal{F}^{(s)}(\hat{u}^{(s)})$ of our model 
is calculated by automatic differentiation~\cite{paszke2019pytorch}.
By using the PINN error in Eq.~(\ref{eq:theta}),
we can meta-learn our model by feeding PDE problems without their solutions.

Algorithm~\ref{alg:train} shows the meta-learning procedures of our model.
The expectations are approximated with the Monte Carlo method
by randomly sampling PDE problems at Lines~2--4, and sampling points at Lines~5 and 6.
The algorithm requires a random generator for governing equation coefficients, domains, boundaries, and boundary conditions,
by which we can specify a class of PDE problems we can solve.
Since our model is differentiable,
we can update parameters $\bm{\Theta}$ using a stochastic gradient descent method~\cite{kingma2014adam}.
By minimizing the expected error with Eq.~(\ref{eq:theta}),
the generalization performance for unseen PDE problems can be improved.
We can generate training data unlimitedly
since the proposed method does not require any observational data.


\begin{algorithm}[t!]
  \centering
  \caption{Meta-learning procedures of our model.}
  \label{alg:train}
  \begin{algorithmic}[1]
    \renewcommand{\algorithmicrequire}{\textbf{Input:}}
    \renewcommand{\algorithmicensure}{\textbf{Output:}}
    \REQUIRE{Point dimension $D$,
      maximum degree of polynomial $C$, and maximum order of derivative $J$ of governing equations,
      number of sampling domain points $N_{\mathrm{f}}$,      
      number of sampling boundary points $N_{\mathrm{g}}$,
     random generator for governing equation coefficients, domains, boundaries, and boundary conditions.}
    \ENSURE{Trained model parameters $\bm{\Theta}$.}
    \WHILE{End condition is satisfied}
    \STATE Randomly generate domain $\Omega^{(s)}$ and boundary $\partial\Omega^{(s)}$.
    \STATE Randomly generate governing equation coefficients $\bm{\alpha}^{(s)}$.
    \STATE Randomly generate boundary condition $g^{(s)}$.
    \STATE Randomly generate $N_{\mathrm{g}}$ points $\vec{X}_{\mathrm{g}}$ on boundary $\partial\Omega^{(s)}$.
    \STATE Randomly generate $N_{\mathrm{f}}$ points $\vec{X}_{\mathrm{f}}=\{\vec{x}_{\mathrm{f}n}\}_{n=1}^{N_{\mathrm{f}}}$
    in domain $\Omega^{(s)}$.
    \STATE Evaluate governing equation error
    $E_{\mathrm{f}} =
    \frac{1}{N_{\mathrm{f}}}
    \sum_{n=1}^{N_{\mathrm{f}}}\parallel 
    \mathcal{F}^{(s)}\left(\hat{u}^{(s)}(\vec{x}_{\mathrm{f}n},\mathcal{F}^{(s)},g^{(s)},\vec{X}_{\mathrm{g}})\right)
    \parallel^{2}$.
    \STATE Evaluate boundary condition error
    $E_{\mathrm{g}} =    
    \frac{1}{N_{\mathrm{g}}}
    \sum_{n=1}^{N_{\mathrm{g}}}
        \parallel 
        g(\vec{x}_{\mathrm{g}n})-
        \hat{u}^{(s)}(\vec{x}_{\mathrm{g}n},\mathcal{F}^{(s)},g^{(s)},\vec{X}_{\mathrm{g}})
        \parallel^{2}$.
        \STATE Update model parameters $\bm{\Theta}$ using gradients of the sum of
        losses $E_{\mathrm{f}}+E_{\mathrm{g}}$ by a stochastic gradient method.
    \ENDWHILE
  \end{algorithmic}
\end{algorithm}

\subsection{Finetuning}

The performance of our trained model on a PDE problem
can be evaluated instantly using the PINN error.
For applications that require low errors,
we can improve the approximation performance
of the predicted solution by our model with finetuning~\cite{huang2022meta}.
Suppose that our model is pretrained using various PDE problems with the procedures described in
Section~\ref{sec:meta-learning}.
First, given a PDE problem,
we calculate its problem representation $\bm{z}^{(s)}$
using pretrained neural networks $\mathrm{NN_{b1}}$, $\mathrm{NN_{b2}}$, and $\mathrm{NN_{z}}$.
Next,
we finetune the parameters of neural network $\mathrm{NN_{u}}$,
and problem representation $\bm{z}^{(s)}$
by minimizing the PINN error while fixing $\mathrm{NN_{b1}}$, $\mathrm{NN_{b2}}$, and $\mathrm{NN_{z}}$ as follows:
\begin{align}
  \hat{\bm{\theta}}^{(s)}_{\mathrm{u}},\hat{\vec{z}}^{(s)}=
  \argmin_{\bm{\theta}_{\mathrm{u}},\vec{z}^{(s)}}&
  \frac{1}{N_{\mathrm{f}}}\sum_{n=1}^{N_{\mathrm{f}}}
  \parallel \mathcal{F}^{(s)}\big(\mathrm{NN}_{\mathrm{u}}(\vec{x}_{\mathrm{f}n},\vec{z}^{(s)};\bm{\theta}_{\mathrm{u}})\big)\parallel^{2}
  \nonumber\\
  &+
  \frac{1}{N_{\mathrm{g}}}\sum_{n=1}^{N_{\mathrm{g}}}
  \parallel g(\vec{x}_{\mathrm{g}n})-\mathrm{NN}_{\mathrm{u}}(\vec{x}_{\mathrm{g}n},\vec{z}^{(s)};\bm{\theta}_{\mathrm{u}})\parallel^{2},
      \label{eq:finetuning}
\end{align}
where $\bm{\theta}_{\mathrm{u}}$ is the parameters in $\mathrm{NN_{u}}$,
$\{\vec{x}_{\mathrm{f}n}\}_{n=1}^{N_{\mathrm{f}}}$ and $\{\vec{x}_{\mathrm{g}n}\}_{n=1}^{N_{\mathrm{g}}}$ are sets
of randomly sampled points in domain $\Omega^{(s)}$ and boundary $\partial\Omega^{(s)}$, respectively.
Here, $\bm{\theta}_{\mathrm{u}}$ is initialized by the pretrained model,
and $\vec{z}^{(s)}$ is initialized by the obtained problem representation in the first step.
Using these initializations, we can incorporate meta-learned knowledge for solving PDE problems.
We do not need to finetune neural networks
$\mathrm{NN_{b1}}$, $\mathrm{NN_{b2}}$, and $\mathrm{NN_{z}}$
by finetuning problem representation $\vec{z}^{(s)}$.
Finetuning the problem representation is effective
since the number of parameters in the problem representation
is much smaller than the number of parameters in these neural networks.

\section{Experiments}
\label{sec:experiments}

\subsection{Settings}
\label{sec:settings}

We evaluated the proposed method using PDE problems 
with point $\vec{x}=(t,x)$ of dimension $D=2$,
maximum degree of the polynomial $C=2$,
and maximum order of the derivatives $J=2$.
In this case, the dimension of coefficient vector $\bm{\alpha}^{(s)}$ was 28.
Each coefficient $\alpha_{c}^{(s)}$ was set at zero with probability 0.75
and otherwise sampled uniformly from $[-1,1]$.
The domains for all PDE problems were identical with $t\in[0,1]$ and $x\in[-1,1]$.
The boundary conditions at $x=-1$ and $x=1$ were fixed at $g^{(s)}(t,-1)=g^{(s)}(t,1)=0$ for all problems.
The boundary conditions at $t=0$ (or initial condition) depended on the problem,
where $g^{(s)}(0,x)=(x-1)(x+1)(r^{(s)}_{1}x^{2}+r^{(s)}_{2}x+r^{(s)}_{3})$,
and $(r^{(s)}_{1},r^{(s)}_{2},r^{(s)}_{3})$ were problem-specific coefficients
sampled uniform randomly from $[-1,1]$.

\subsection{Compared methods}

We compared the proposed method with
the neural process (NP)~\cite{garnelo2018conditional},
model-agnostic meta-learning (MAML)~\cite{finn2017model},
and multi-task learning (MT).
All methods are trained by minimizing the expected PINN loss.
The NP uses a neural network that takes boundary condition information and a point as input,
and outputs a predicted solution at the point.
NP corresponds to the proposed method without governing equation information.
MAML and MT use neural networks that take a point as input,
and outputs a predicted solution at the point, as with the PINN.
In MAML, the initial values of the model parameters are estimated
such that the expected PINN loss is minimized when adapted by gradient descent.
In MT, the model parameters are estimated by minimizing the expected PINN loss
without problem adaptation.

\subsection{Implementation}

We used the same architecture of neural networks for all methods when applicable.
For $\mathrm{NN_{b1}}$ and $\mathrm{NN_{b2}}$,
we used four-layered feed-forward neural networks with 256 hidden and output units.
For the input of $\mathrm{NN_{b1}}$,
boundary point $\vec{x}_{gn}$ and it solution
$g^(s)(\vec{x}_{gn})$ were concatenated.
In $\mathrm{NN_{z}}$,
coefficients $\bm{\alpha}^{(s)}$ were transformed by
a four-layered feed-forward neural network with 256 hidden and output units,
its output was concatenated with boundary representation $\bm{\beta}^{(s)}$,
and then transformed into problem representation $\vec{z}^{(s)}$
by a linear layer with 256 output units.
In $\mathrm{NN_{u}}$,
point $\vec{x}$ was transformed by a linear layer with 256 output units,
its output was concatenated with problem representation $\vec{z}^{(s)}$,
and then transformed by a five-layered neural network with 256 hidden units.
All hidden layers had residual connections.
We used sinusoidal activation functions~\cite{sitzmann2020implicit} in $\mathrm{NN_{u}}$,
and rectified linear activation functions in the other neural networks.
With NP, the output of $\mathrm{NN_{b2}}$ was used as problem representation $\vec{z}^{(s)}$.
In $\mathrm{NN_{u}}$ of MAML and MT,
there was no concatenation with problem representation $\vec{z}^{(s)}$.
In MAML, for adapting to each PDE problem,
we used one step of the gradient descent with learning rate $10^{-2}$,
where a computational graph with more than one steps for the adaptation could not stored in memory.

In all methods,
PDE problems were randomly generated in each meta-learning iteration
based on the settings described in the first paragraph of Section~\ref{sec:experiments}.
For evaluating the governing equation error at Line 7 in Algorithm~\ref{alg:train},
we used $N_{\mathrm{f}}=50$ uniform ramdomly sampled points in the domain.
For evaluating the boundary condition error at Line 8,
we used $N_{\mathrm{g}}=100$ randomly sampled points on the boundary.
In particular, we used 50 initial points at $t=0$,
and 25 boundary points at $x=-1$ and $x=1$, respectively.
The number of epochs was 30,000, where 
each epoch consisted of 9,800 PDE problems.
We optimized using Adam~\cite{kingma2014adam} with initial learning rate $10^{-3}$,
which was decayed by half for every 5,000 epochs.
In a mini-batch, we used 512 PDE problems.
We implemented all methods with PyTorch~\cite{paszke2019pytorch}.

For the evaluation measurement, we used the PINN error in Eq~(\ref{eq:pinn_loss}),
the governing equation (GE) error,
which is the first term in Eq~(\ref{eq:pinn_loss}),
and boundary condition (BC) error,
which is the second term in Eq~(\ref{eq:pinn_loss}).
The errors were evaluated on 100 PDE problems,
where we used $N_{\mathrm{f}}=10000$ sampled points in the domain,
and $N_{\mathrm{g}}=100$ sampled points on the boundary.




\subsection{Results}

Table~\ref{tab:error} shows the average error.
The proposed method achieved the best performance
in terms of not only the total PINN error
but also the governing equation and boundary condition errors.
The error by NP was higher than that by the proposed method.
This result indicates the effectiveness of including governing equation information in the proposed method.
Since MAML could not perform many gradient descent steps for problem adaptation,
the error was high.
On the other hand,
the proposed method approximates problem adaptation with neural networks,
and the neural networks are trained with many randomly generated problems,
it achieved a low error.
Since MT does not output problem-specific solutions, its error was high.



\begin{table}[t!]
  \centering
  \caption{Average PINN errors, governing equation (GE) errors, and boundary condition (BC) errors with their standard errors. Values in bold typeface are not statistically significantly different at the 5\% level from the best performing method in each row according to a paired t-test.}
  \label{tab:error}
  \begin{tabular}{lrrrr}
    \hline
    & Ours & NP & MAML & MT \\
    \hline
        PINN error & {\bf 0.034 $\pm$ 0.004} & 0.157 $\pm$ 0.026 & 0.131 $\pm$ 0.021 & 0.201 $\pm$ 0.025\\
        GE error & {\bf 0.009 $\pm$ 0.001} & 0.089 $\pm$ 0.025 & 0.062 $\pm$ 0.018 & 0.082 $\pm$ 0.023\\
        BC error & {\bf 0.024 $\pm$ 0.003} & 0.068 $\pm$ 0.005 & 0.069 $\pm$ 0.006 & 0.119 $\pm$ 0.009\\
    \hline
  \end{tabular}
\end{table}

\begin{figure}[t!]
  \centering
\begin{tabular}{cccc}
  Solution & Ours & NP & MAML \\
\includegraphics[width=9.5em]{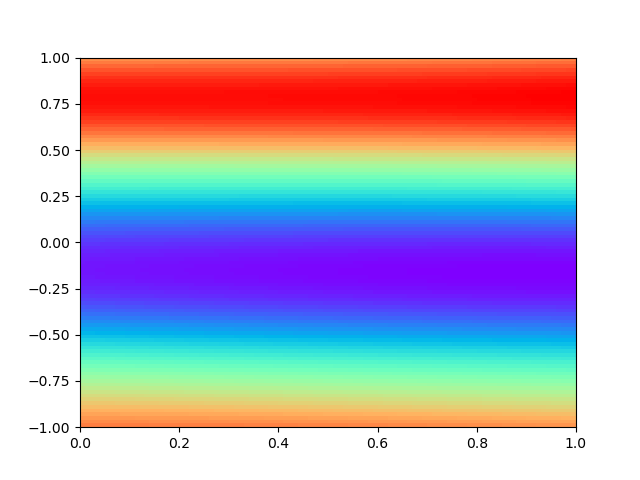} & \includegraphics[width=9.5em]{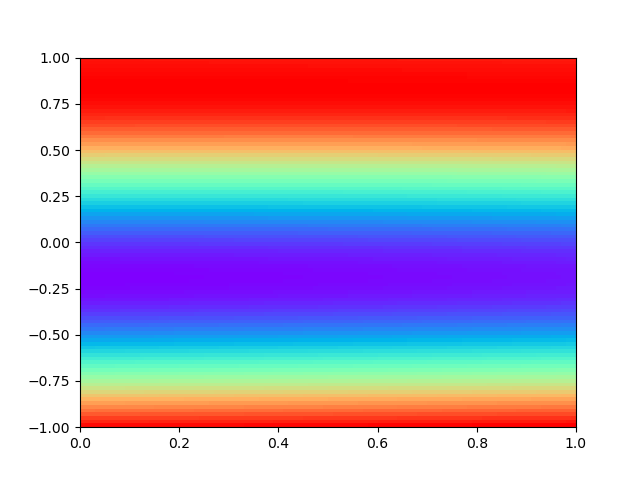} & \includegraphics[width=9.5em]{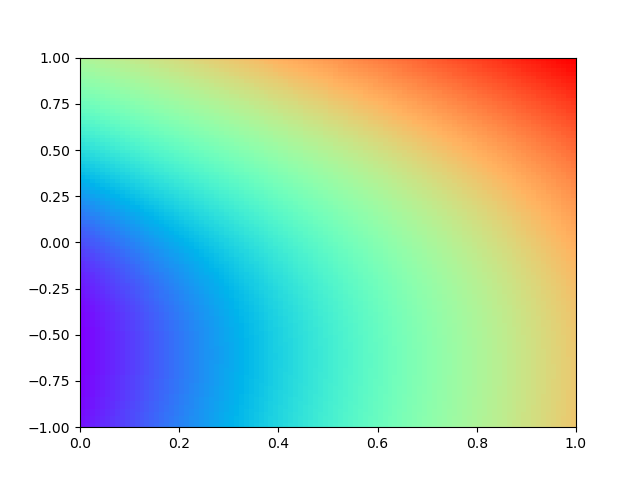} & \includegraphics[width=9.5em]{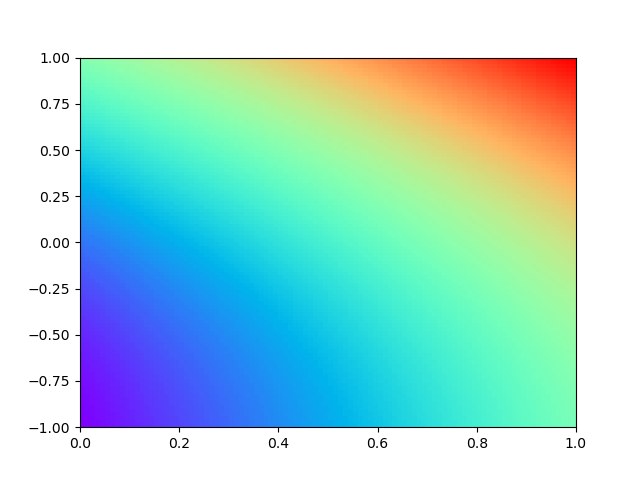} \\
0.00009 & 0.00540 & 0.04720 & 0.04831 \\
\end{tabular}
\\
(a) {\small $-0.416u_{tx}u_{xx}-0.828u_{tt}+0.685u_{t}-0.726u_{t}u_{tx}+0.893u_{t}u_{x}+0.993u_{t}^{2}-0.827uu_{t}+0.388u^{2}=0$}\\
\begin{tabular}{cccc}
\includegraphics[width=9.5em]{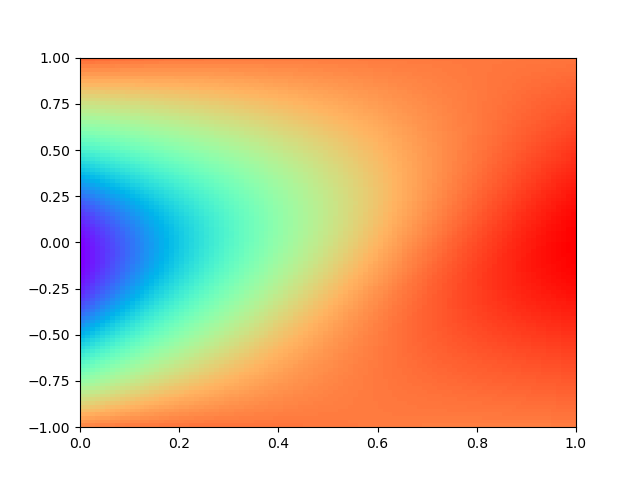} & \includegraphics[width=9.5em]{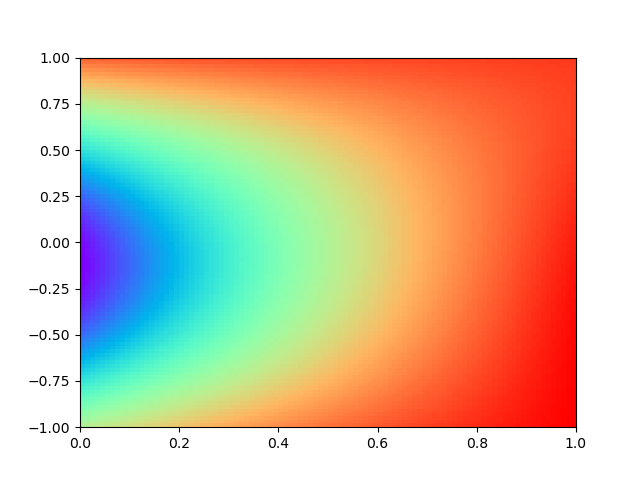} & \includegraphics[width=9.5em]{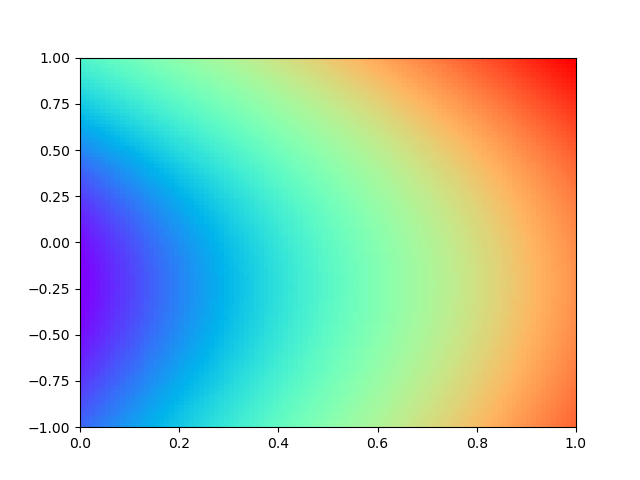} & \includegraphics[width=9.5em]{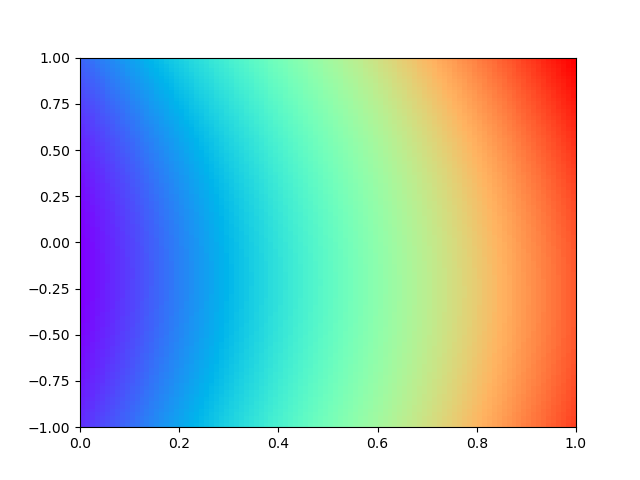} \\
0.00005 & 0.00416 & 0.03796 & 0.03994 \\
\end{tabular}
\\
(b) {\small $-0.305u_{xx}-0.189u_{tx}u_{xx}-0.511u_{tt}^{2}-0.936u_{t}u_{tt}=0$}
  \\
\begin{tabular}{cccc}
\includegraphics[width=9.5em]{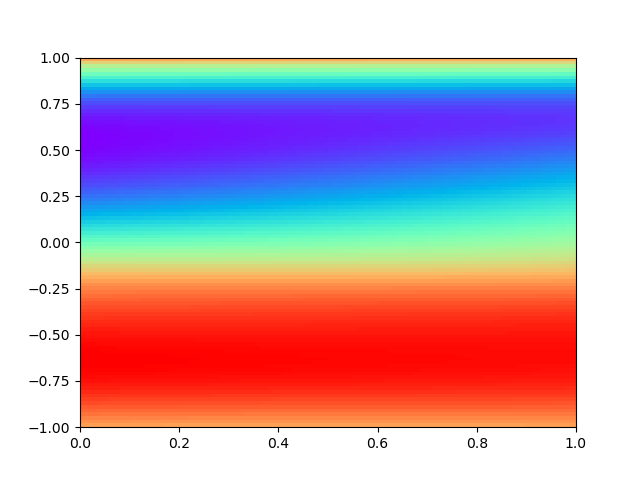} & \includegraphics[width=9.5em]{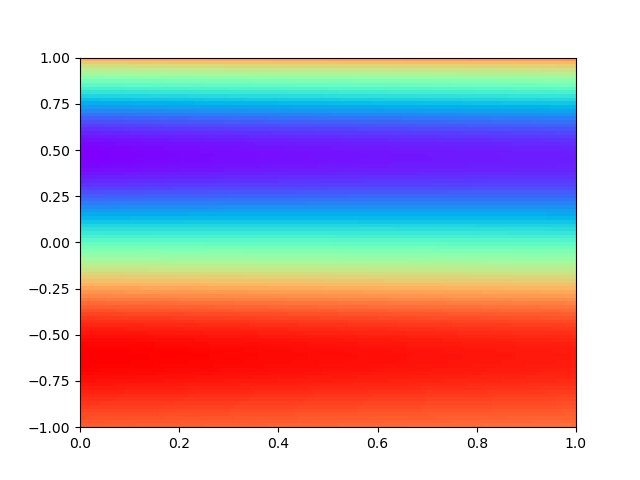} & \includegraphics[width=9.5em]{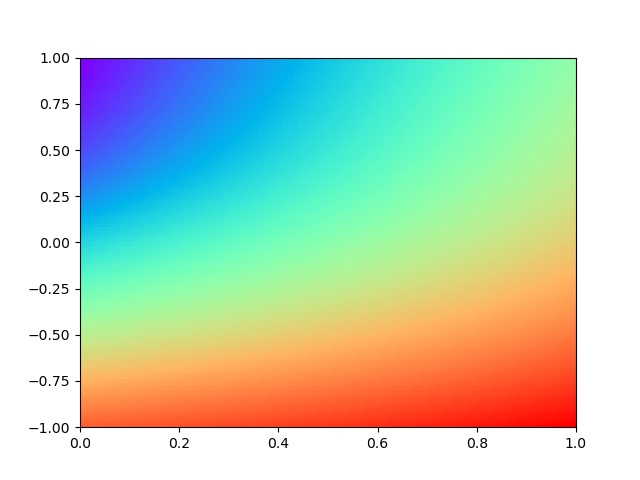} & \includegraphics[width=9.5em]{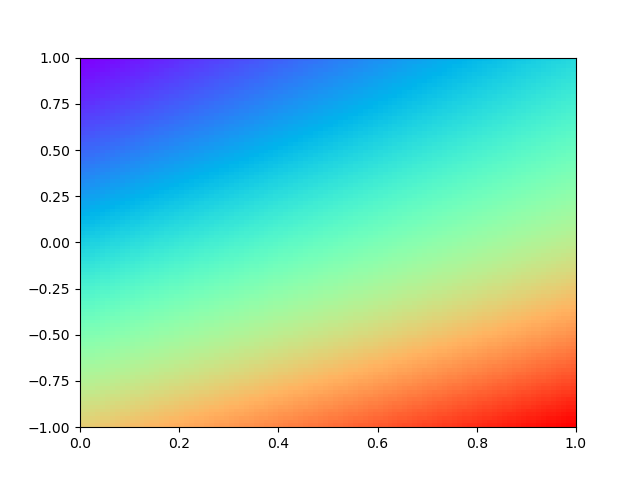} \\
0.00020 & 0.00546 & 0.02825 & 0.02833 \\
\end{tabular}
\\
(c) {\small $-0.222u_{t}-0.821u_{t}u_{xx}+0.112u_{t}u_{tt}-0.655u_{t}u_{x}+0.256uu_{x}+0.798uu_{t}+0.525u^{2}=0$}\\
\begin{tabular}{cccc}
\includegraphics[width=9.5em]{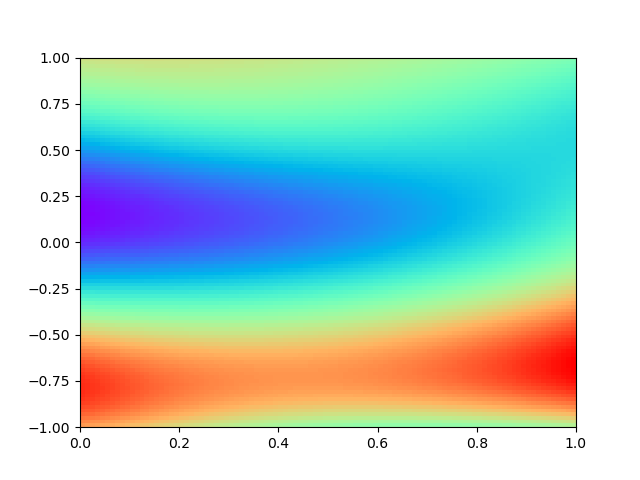} & \includegraphics[width=9.5em]{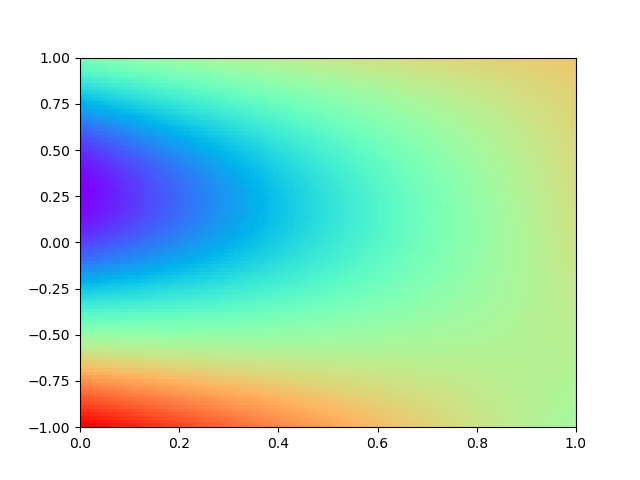} & \includegraphics[width=9.5em]{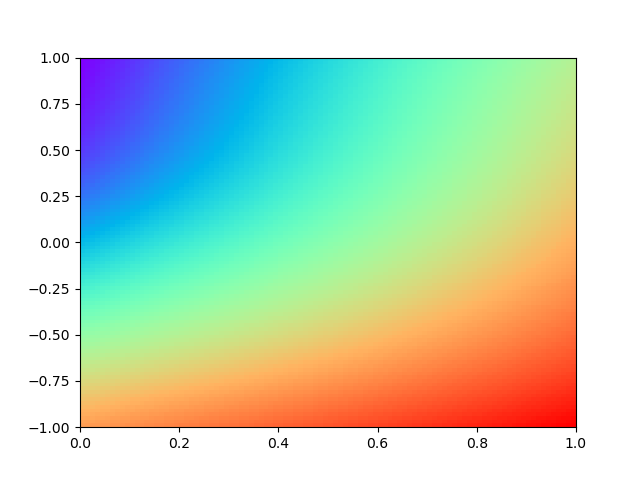} & \includegraphics[width=9.5em]{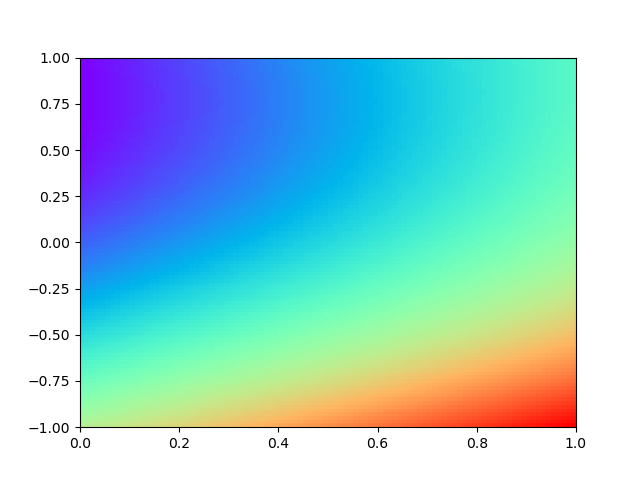} \\
0.00083 & 0.00394 & 0.00616 & 0.00613 \\
\end{tabular}
\\
(d) {\small $0.855u_{tt}u_{tx}-0.403u_{x}-0.403u^{2}=0$}\\
\begin{tabular}{cccc}
\includegraphics[width=9.5em]{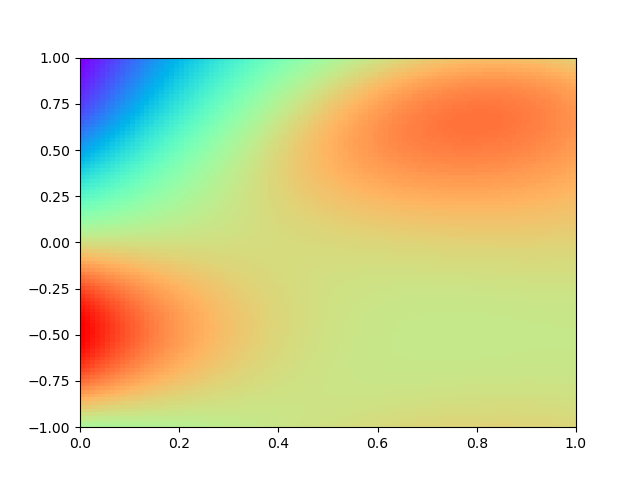} & \includegraphics[width=9.5em]{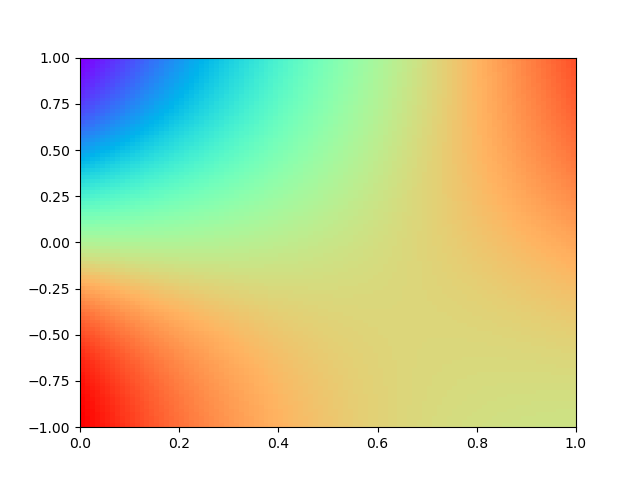} & \includegraphics[width=9.5em]{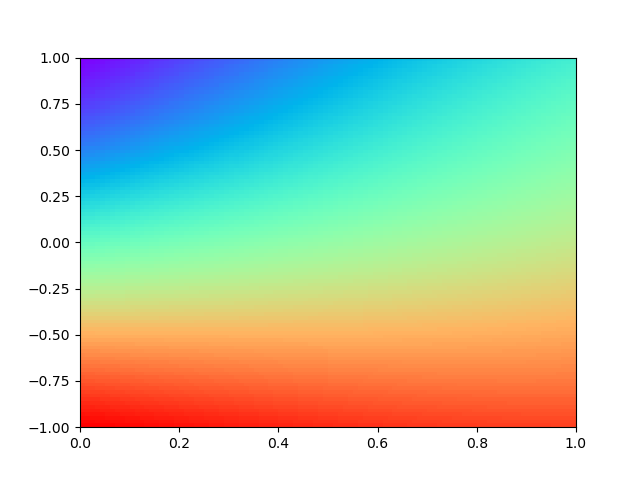} & \includegraphics[width=9.5em]{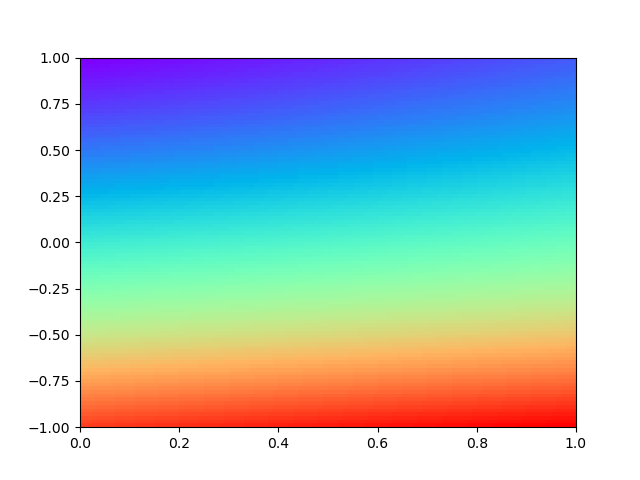} \\
0.00847 & 0.01894 & 0.02609 & 0.02487 \\
\end{tabular}
\\
(e) {\scriptsize $0.107-0.805u_{xx}-0.962u_{tt}^{2}-0.469u_{x}u_{xx}+0.355u_{x}u_{tx}+0.920u_{x}u_{tt}-0.586u_{t}u_{tx}-0.902uu_{tt}-0.306uu_{t}=0$}\\
\begin{tabular}{cccc}
\includegraphics[width=9.5em]{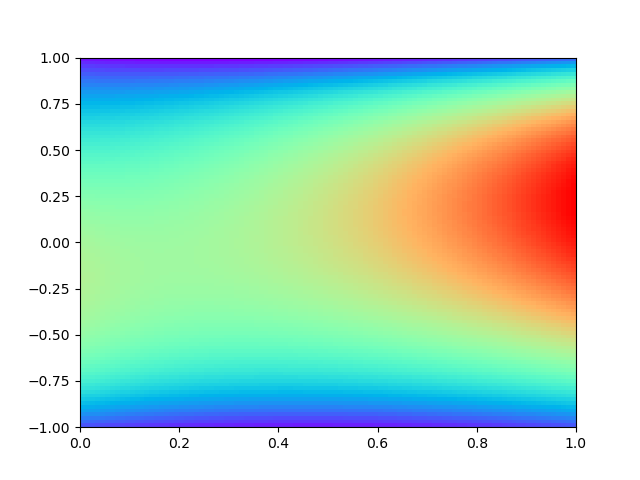} & \includegraphics[width=9.5em]{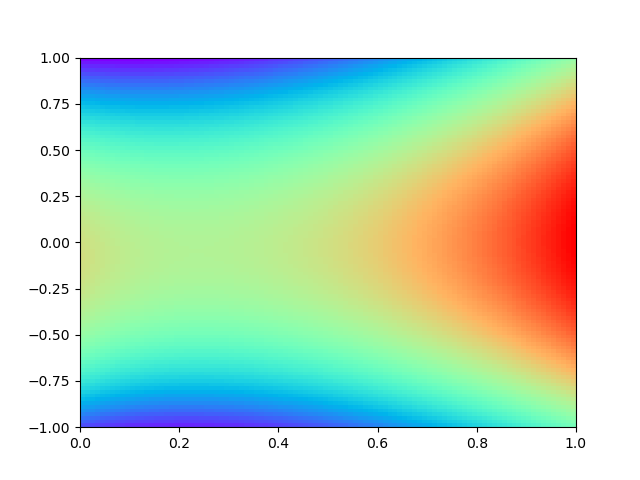} & \includegraphics[width=9.5em]{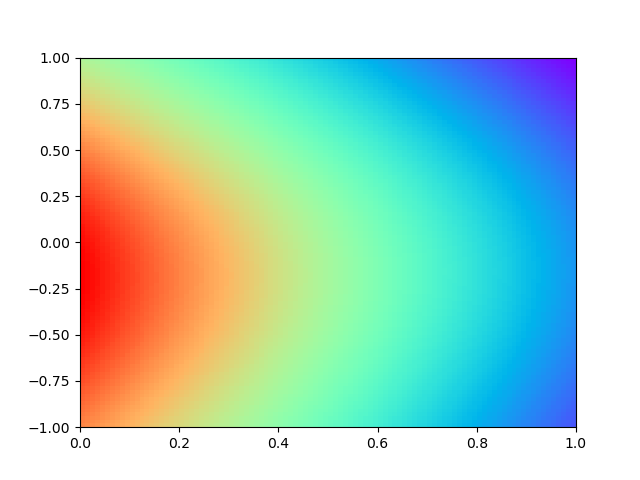} & \includegraphics[width=9.5em]{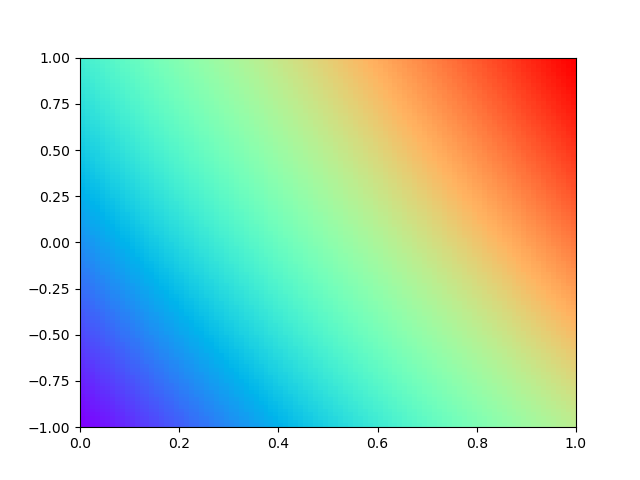} \\
0.00071 & 0.01034 & 0.83029 & 0.53358 \\
\end{tabular}
\\
(f) {\scriptsize $0.870-0.562u_{tx}+0.932u_{tt}u_{xx}-0.674u_{t}+0.635u_{t}u_{tx}-0.709u_{t}u_{tt}-0.691u_{t}u_{x}+0.495u_{t}^{2}+0.463uu_{xx}=0$}\\
\caption{Approximated solutions by the proposed method (Ours), NP, and MAML for six PDE problems. The equations show the governing equations. The value below the figure shows the PINN error.
  The left column (Solution) shows the solutions by PINNs with 30,000 training epochs for each problem.}
\label{fig:solution}
\end{figure}

Figure~\ref{fig:solution} shows examples of approximated solutions by the proposed method, NP, and MAML.
The proposed method's outputs were more similar to the solution than the other methods.
The proposed method predicted different patterns of solutions depending on the given PDE problems
without any problem-specific parameter updates.

Table~\ref{tab:train_time} shows the computation time in hours for meta-learning of 30,000 epochs
using computers with Tesla v100 GPU, Xeon Gold 6148 2.40GHz, and 512GB memory.
Since MAML requires gradient descent steps for problem adaptation
in each meta-learning iteration, it took a longer time.
MT does not have neural networks for calculating problem representations,
which resulted in a shorter meta-training time.

\begin{table}[t!]
  \centering
  \caption{Computation time in hours for meta-learning.}
  \label{tab:train_time}
  \begin{tabular}{rrrr}
    \hline
    Ours & NP & MAML & MT \\
    \hline
    51.1 & 50.2 & 209.2 & 37.2\\
    \hline
  \end{tabular}
\end{table}

Table~\ref{tab:inference_time} shows the average computation time in seconds for predicting solutions of
at 10,000 points in the domain and 100 points on the boundary
using computers with Xeon Gold 6130 CPU 2.10GHz and 256GB memory without GPU.
The proposed method can predict solutions of newly given PDE problems in about a second.
The solutions in Figure~\ref{fig:solution} were calculated by the PINN with 30,000 training epochs,
which took 27.7 hours per PDE problem on average.

\begin{table}[t!]
  \centering
  \caption{Average computation times in seconds for predicting solutions at 10,100 points, and their standard errors.}
  \label{tab:inference_time}
  \begin{tabular}{rrrr}
    \hline
    Ours & NP & MAML & MT \\
    \hline
    1.457 $\pm$ 0.015 & 1.468 $\pm$ 0.013 & 7.030 $\pm$ 0.067 & 1.287 $\pm$ 0.016\\    
    \hline
  \end{tabular}
\end{table}

In the proposed method, we randomly generate PDE problems for each meta-learning iteration
as described in Algorithm~\ref{alg:train}.
Table~\ref{tab:n_task} shows the PINN errors 
when we used a fixed number of PDE problems $\{1000,2000,4000,6000,8000\}$
for meta-training in the proposed method.
When our model is trained with a fixed set of PDE problems,
even when many (8,000) PDE problems were used, the performance was low.
This result indicates the importance of using a wide variety of PDE problems for meta-learning by random sampling.
In our model, since we use neural networks to obtain problem representations given PDE problems,
we can use different PDE problems in each iteration for meta-learning.
On the other hand, implicit encoding 
of problem representation
in Meta-Auto-Decoder~\cite{huang2022meta}
requires to infer the representation for each meta-learning PDE problem,
which prohibits to use randomly generated problems for each iteration.

\begin{table}[t!]
  \centering
  \caption{Average PINN errors with a fixed number of PDE problems for meta-learning, and their standard errors.
    $\infty$ represents that PDE problems are randomly generated for each meta-learning iteration.}
  \label{tab:n_task}
  \begin{small}
  {\tabcolsep=0.3em
  \begin{tabular}{lrrrrrrr}
    \hline
    \#tasks & 1000 & 2000 & 4000 & 6000 & 8000 & $\infty$ \\
    \hline
    PINN Error & 93.217 $\pm$ 87.191 & 4.144 $\pm$ 3.138 & 0.731 $\pm$ 0.150 & 0.658 $\pm$ 0.106 & 1.110 $\pm$ 0.500 & {\bf 0.034 $\pm$ 0.004}\\
    \hline
  \end{tabular}}
  \end{small}
\end{table}

\begin{figure}[t!]
  \centering
  \includegraphics[width=18em]{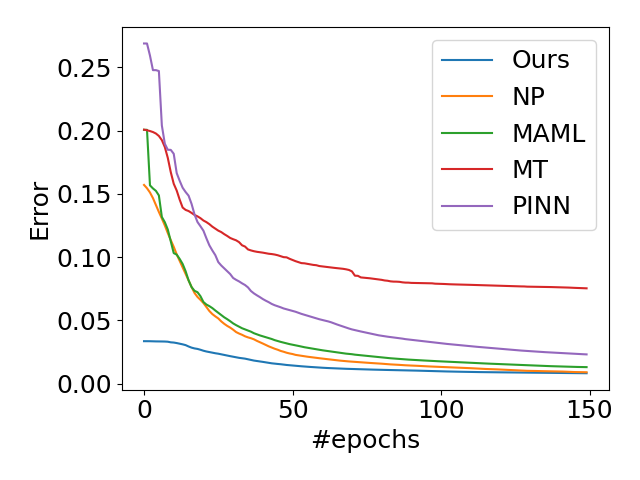}
  \caption{Average PINN errors when finetuned. The horizontal axis shows the number of epochs for finetuning.}
  \label{fig:finetune}
\end{figure}

Figure~\ref{fig:finetune} shows the PINN errors when finetuned for each PDE problem.
Here, PINN represents the physics-informed neural network
with the same architecture as MAML and MT,
where its parameters were initialized randomly.
The proposed method achieved a low error with finetuning.
Since NP and MAML were meta-learning methods,
their errors were lower than PINN but higher than the proposed method.
The error by MT without finetuning was lower than PINN,
but after a few dozen finetuning epochs, PINN outperformed MT.
Since MT does not have problem-specific parameters,
it is difficult for MT to adapt to various PDE problems flexibly.
In contrast, since the proposed method has a problem representation to be finetuned,
it outperformed MT.
The performance by the proposed method without any finetuning
was comparable to PINN with 94 finetuning epochs, which took 312 seconds on average.
      
\section{Conclusion}

In this paper, we proposed a meta-learning method for solving PDE problems
based on physics-informed neural networks.
Although we believe that our work is an important step
for accumulating reusable knowledge of efficiently solving PDE problems,
we must extend our approach in several directions.
First, we plan to enlarge the class of governing equations
by encoding their mathematical expressions using neural networks~\cite{kamiennyend2022}.
Second, we will use our method for discovering governing equations
from observational data by estimating
the coefficients of governing equations in our model.

\bibliographystyle{abbrv}
\bibliography{arxiv_pinn}

\begin{thebibliography}{10}

\bibitem{courant2008methods}
R.~Courant and D.~Hilbert.
\newblock {\em Methods of mathematical physics: partial differential
  equations}.
\newblock John Wiley \& Sons, 2008.

\bibitem{de2021hyperpinn}
F.~de~Avila Belbute-Peres, Y.-f. Chen, and F.~Sha.
\newblock Hyper{PINN}: Learning parameterized differential equations with
  physics-informed hypernetworks.
\newblock In {\em The Symbiosis of Deep Learning and Differential Equations},
  2021.

\bibitem{dong2023survey}
Q.~Dong, L.~Li, D.~Dai, C.~Zheng, Z.~Wu, B.~Chang, X.~Sun, J.~Xu, L.~Li, and
  Z.~Sui.
\newblock A survey on in-context learning.
\newblock {\em arXiv preprint arXiv:2301.00234}, 2023.

\bibitem{duffy2013finite}
D.~J. Duffy.
\newblock {\em Finite difference methods in financial engineering: a partial
  differential equation approach}.
\newblock John Wiley \& Sons, 2013.

\bibitem{finn2017model}
C.~Finn, P.~Abbeel, and S.~Levine.
\newblock Model-agnostic meta-learning for fast adaptation of deep networks.
\newblock In {\em International Conference on Machine Learning}, pages
  1126--1135, 2017.

\bibitem{garnelo2018conditional}
M.~Garnelo, D.~Rosenbaum, C.~Maddison, T.~Ramalho, D.~Saxton, M.~Shanahan,
  Y.~W. Teh, D.~Rezende, and S.~A. Eslami.
\newblock Conditional neural processes.
\newblock In {\em International conference on machine learning}, pages
  1704--1713, 2018.

\bibitem{huang2022meta}
X.~Huang, Z.~Ye, H.~Liu, S.~Ji, Z.~Wang, K.~Yang, Y.~Li, M.~Wang, H.~Chu,
  F.~Yu, et~al.
\newblock Meta-auto-decoder for solving parametric partial differential
  equations.
\newblock {\em Advances in Neural Information Processing Systems},
  35:23426--23438, 2022.

\bibitem{jones2009differential}
D.~S. Jones, M.~Plank, and B.~D. Sleeman.
\newblock {\em Differential equations and mathematical biology}.
\newblock CRC press, 2009.

\bibitem{kamiennyend2022}
P.-A. Kamienny, S.~d'Ascoli, G.~Lample, and F.~Charton.
\newblock End-to-end symbolic regression with transformers.
\newblock {\em Advances in Neural Information Processing Systems}, 35, 2022.

\bibitem{kaplan2020scaling}
J.~Kaplan, S.~McCandlish, T.~Henighan, T.~B. Brown, B.~Chess, R.~Child,
  S.~Gray, A.~Radford, J.~Wu, and D.~Amodei.
\newblock Scaling laws for neural language models.
\newblock {\em arXiv preprint arXiv:2001.08361}, 2020.

\bibitem{karniadakis2021physics}
G.~E. Karniadakis, I.~G. Kevrekidis, L.~Lu, P.~Perdikaris, S.~Wang, and
  L.~Yang.
\newblock Physics-informed machine learning.
\newblock {\em Nature Reviews Physics}, 3(6):422--440, 2021.

\bibitem{kingma2014adam}
D.~P. Kingma and J.~Ba.
\newblock {ADAM}: {A} method for stochastic optimization.
\newblock In {\em International Conference on Learning Representations}, 2015.

\bibitem{lin1988mathematics}
C.-C. Lin and L.~A. Segel.
\newblock {\em Mathematics applied to deterministic problems in the natural
  sciences}.
\newblock SIAM, 1988.

\bibitem{pan2023neural}
S.~Pan, S.~L. Brunton, and J.~N. Kutz.
\newblock Neural implicit flow: a mesh-agnostic dimensionality reduction
  paradigm of spatio-temporal data.
\newblock {\em Journal of Machine Learning Research}, 24(41):1--60, 2023.

\bibitem{paszke2019pytorch}
A.~Paszke, S.~Gross, F.~Massa, A.~Lerer, J.~Bradbury, G.~Chanan, T.~Killeen,
  Z.~Lin, N.~Gimelshein, L.~Antiga, et~al.
\newblock Pytorch: An imperative style, high-performance deep learning library.
\newblock {\em Advances in Neural Information Processing Systems}, 32, 2019.

\bibitem{qin2022meta}
T.~Qin, A.~Beatson, D.~Oktay, N.~McGreivy, and R.~P. Adams.
\newblock Meta-{PDE}: Learning to solve {PDE}s quickly without a mesh.
\newblock {\em arXiv preprint arXiv:2211.01604}, 2022.

\bibitem{raissi2019physics}
M.~Raissi, P.~Perdikaris, and G.~E. Karniadakis.
\newblock Physics-informed neural networks: A deep learning framework for
  solving forward and inverse problems involving nonlinear partial differential
  equations.
\newblock {\em Journal of Computational Physics}, 378:686--707, 2019.

\bibitem{sitzmann2020implicit}
V.~Sitzmann, J.~Martel, A.~Bergman, D.~Lindell, and G.~Wetzstein.
\newblock Implicit neural representations with periodic activation functions.
\newblock {\em Advances in Neural Information Processing Systems},
  33:7462--7473, 2020.

\bibitem{turner1956stiffness}
M.~J. Turner, R.~W. Clough, H.~C. Martin, and L.~Topp.
\newblock Stiffness and deflection analysis of complex structures.
\newblock {\em Journal of the Aeronautical Sciences}, 23(9):805--823, 1956.

\bibitem{xu2020metafun}
J.~Xu, J.-F. Ton, H.~Kim, A.~Kosiorek, and Y.~W. Teh.
\newblock Metafun: Meta-learning with iterative functional updates.
\newblock In {\em International Conference on Machine Learning}, pages
  10617--10627, 2020.

\bibitem{zaheer2017deep}
M.~Zaheer, S.~Kottur, S.~Ravanbakhsh, B.~Poczos, R.~R. Salakhutdinov, and A.~J.
  Smola.
\newblock Deep sets.
\newblock {\em Advances in Neural Information Processing Systems}, 30, 2017.

\end{thebibliography}

\end{document}